\documentclass{wscpaperproc}
\usepackage{latexsym}
\usepackage{graphicx}
\usepackage{mathptmx}
\usepackage[T1]{fontenc}

\usepackage{amsmath}
\usepackage{amssymb}
\usepackage{appendix}
\usepackage{booktabs}
\usepackage[shortlabels]{enumitem}
\usepackage{float}
\usepackage{subcaption}
\usepackage{tabulary}
\usepackage{xfrac}
\usepackage{xurl}
\usepackage[pdftex,colorlinks=true,urlcolor=blue,citecolor=black,anchorcolor=black,linkcolor=black]{hyperref}

\begin{document}

%***************************************************************************
% AUTHOR: AUTHOR NAMES GO HERE
% FORMAT AUTHORS NAMES Like: Author1, Author2 and Author3 (last names)
%
%		You need to change the author listing below!
%               Please list ALL authors using last name only, separate by a comma except
%               for the last author, separate with "and"
%

% setting up general page style
\pagestyle{fancyplain}

% setting up page style of first page
\thispagestyle{plain}
\firstPageHead{}

% setting up running header (authors) of subsequent pages
\chead{\fancyplain{}{\itshape Chen, Carley, Fang, and Sadeh}}

% setting up seperation parameters
%\headsep=72pt
\rhead{}
\cfoot{}
\renewcommand{\headrulewidth}{0pt} % (renewcommand needed in fancyhdr to remove top decorative line)
%\headrulewidth=0pt  % ("setlength" needed in fancyheading to remove top decorative line)

%%%%%%%%%%%%%%%%%%%%%%%%%%%%%%%%%%%%%%%%%%%%%%%%%%%%%%%%%%%%%%%%%%%%%%%%%%%%%%
%                                                                            %
%     THESE COMMANDS ARE REQUIRED TO WORK WITH WSC.BST TO MAKE BIBLIO     %
%                                                                            %
%%%%%%%%%%%%%%%%%%%%%%%%%%%%%%%%%%%%%%%%%%%%%%%%%%%%%%%%%%%%%%%%%%%%%%%%%%%%%%
\makeatletter
\let\@internalcite\cite
\def\cite{\def\@citeseppen{-1000}%
    \def\@cite##1##2{(##1\if@tempswa , ##2\fi)}%
    \def\citeauthoryear##1##2##3{##1 ##3}\@internalcite}
\def\citeNP{\def\@citeseppen{-1000}%
    \def\@cite##1##2{##1\if@tempswa , ##2\fi}%
    \def\citeauthoryear##1##2##3{##1 ##3}\@internalcite}
\def\citeN{\def\@citeseppen{-1000}%
%  Pierre L'Ecuyer's fix for multiple cite bug
%  Added by Paul J Sanchez on 4 October 2001
%   \def\@cite##1##2{##1\if@tempswa , ##2)\else{)}\fi}%
%   \def\citeauthoryear##1##2##3{##1 (##3}\@citedata}
    \def\@cite##1##2{##1\if@tempswa, ##2)\else{}\fi}%
    \def\citeauthoryear##1##2##3{##1 (##3)}\@citedata}
\def\citeA{\def\@citeseppen{-1000}%
    \def\@cite##1##2{(##1\if@tempswa , ##2\fi)}%
    \def\citeauthoryear##1##2##3{##1}\@internalcite}
\def\citeANP{\def\@citeseppen{-1000}%
    \def\@cite##1##2{##1\if@tempswa , ##2\fi}%
    \def\citeauthoryear##1##2##3{##1}\@internalcite}
\def\shortcite{\def\@citeseppen{-1000}%
    \def\@cite##1##2{(##1\if@tempswa , ##2\fi)}%
    \def\citeauthoryear##1##2##3{##2 ##3}\@internalcite}
\def\shortciteNP{\def\@citeseppen{-1000}%
    \def\@cite##1##2{##1\if@tempswa , ##2\fi}%
    \def\citeauthoryear##1##2##3{##2 ##3}\@internalcite}
\def\shortciteN{\def\@citeseppen{-1000}%
%  Pierre L'Ecuyer's fix for multiple cite bug
%  Added by Paul J Sanchez on 2 September 2002
%  should have caught this last year...
%   \def\@cite##1##2{##1\if@tempswa , ##2)\else{)}\fi}%
%   \def\citeauthoryear##1##2##3{##2 (##3}\@citedata}
% Shane G. Henderson fix for extra right bracket at end of optional material June 8, 2005
%    \def\@cite##1##2{##1\if@tempswa, ##2)\else{}\fi}%
    \def\@cite##1##2{##1\if@tempswa, ##2\else{}\fi}%
    \def\citeauthoryear##1##2##3{##2 (##3)}\@citedata}
\def\shortciteA{\def\@citeseppen{-1000}%
    \def\@cite##1##2{(##1\if@tempswa , ##2\fi)}%
    \def\citeauthoryear##1##2##3{##2}\@internalcite}
\def\shortciteANP{\def\@citeseppen{-1000}%
    \def\@cite##1##2{##1\if@tempswa , ##2\fi}%
    \def\citeauthoryear##1##2##3{##2}\@internalcite}
\def\citeyear{\def\@citeseppen{-1000}%
    \def\@cite##1##2{(##1\if@tempswa , ##2\fi)}%
    \def\citeauthoryear##1##2##3{##3}\@citedata}
\def\citeyearNP{\def\@citeseppen{-1000}%
    \def\@cite##1##2{##1\if@tempswa , ##2\fi}%
    \def\citeauthoryear##1##2##3{##3}\@citedata}
%
% \@citedata and \@citedatax:
%
% Place commas in-between citations in the same \citeyear, \citeyearNP,
% \citeN, or \shortciteN command.
% Use something like \citeN{ref1,ref2,ref3} and \citeN{ref4} for a list.
%
\def\@citedata{%
    \@ifnextchar [{\@tempswatrue\@citedatax}%
                  {\@tempswafalse\@citedatax[]}%
}

\def\@citedatax[#1]#2{%
\if@filesw\immediate\write\@auxout{\string\citation{#2}}\fi%
  \def\@citea{}\@cite{\@for\@citeb:=#2\do%
    {\@citea\def\@citea{, }\@ifundefined% by Young
       {b@\@citeb}{{\bf ?}%
       \@warning{Citation `\@citeb' on page \thepage \space undefined}}%
{\csname b@\@citeb\endcsname}}}{#1}}%

% don't box citations, separate with ; and a space
% also, make the penalty between citations negative: a good place to break.
%
\def\@citex[#1]#2{%
\if@filesw\immediate\write\@auxout{\string\citation{#2}}\fi%
  \def\@citea{}\@cite{\@for\@citeb:=#2\do%
    {\@citea\def\@citea{; }\@ifundefined% by Young
       {b@\@citeb}{{\bf ?}%
       \@warning{Citation `\@citeb' on page \thepage \space undefined}}%
{\csname b@\@citeb\endcsname}}}{#1}}%

% (from apalike.sty)
% No labels in the bibliography.
%
\def\@biblabel#1{}
\makeatother

%\newlength{\bibhang}
%\setlength{\bibhang}{2em}

% Indent second and subsequent lines of bibliographic entries. Taken
% from openbib.sty: \newblock is set to {}.
% \renewcommand{\refname}{REFERENCES}

\newdimen\bibindent
\bibindent=0.0em
% SEC: was \def\thebibliography#1{\section*{\refname\@mkboth
% SEC: was   {\uppercase{\refname}}{\uppercase{\refname}}}\list
\def\thebibliography#1{\section*{\refname}\list
   {}{\settowidth\labelwidth{[#1]}
   \leftmargin\parindent
   \itemindent -\parindent
   \listparindent \itemindent
   \itemsep 0pt
   \parsep 0pt}
   \def\newblock{}
   \sloppy
   \sfcode`\.=1000\relax}

           % Set up BiBTeX macros

% needed to make the tex document look more like the word counterpart :-(
\setlength{\baselineskip}{12.7pt}

% AUTHOR: Enter the title, all letters in upper case
\title{\uppercase{Purpose in the Machine: Do Traffic Simulators Produce Distributionally Equivalent Outcomes for Reinforcement Learning Applications?}}

% AUTHOR: Enter the authors of the article, see end of the example document for further examples
\author{
Rex Chen\\
Kathleen M. Carley\\
Fei Fang\\
Norman Sadeh\\[12pt]
School of Computer Science\\
Carnegie Mellon University\\
4665 Forbes Avenue\\
Pittsburgh, PA 15213, USA\\
}

\maketitle

\section*{Abstract}
Traffic simulators are used to generate data for learning in intelligent transportation systems (ITSs). A key question is to what extent their modelling assumptions affect the capabilities of ITSs to adapt to various scenarios when deployed in the real world. This work focuses on two simulators commonly used to train reinforcement learning (RL) agents for traffic applications, CityFlow and SUMO. A controlled virtual experiment varying driver behavior and simulation scale finds evidence against distributional equivalence in RL-relevant measures from these simulators, with the root mean squared error and KL divergence being significantly greater than 0 for all assessed measures. While granular real-world validation generally remains infeasible, these findings suggest that traffic simulators are not a deus ex machina for RL training: understanding the impacts of inter-simulator differences is necessary to train and deploy RL-based ITSs.

\section{Introduction}
\label{sec:introduction}
Transportation efficiency is becoming an increasingly critical challenge due to continual growth in the volume of people and objects that need to be transported. The \textit{2021 Urban Mobility Report} \shortcite{Schrank2021} projected that, while the COVID-19 pandemic alleviated congestion, traffic levels in the US will quickly rebound in areas with expanding populations and job markets to produce the most rapid congestion growth since 1982. The increased traffic will stress existing infrastructure and result in social, economic, and environmental costs \shortcite{Schrank2021}, thus making the development and deployment of intelligent transportation systems (ITSs) a critical priority. At the same time, advances in computational algorithms and roadway infrastructure made in response to these challenges provide opportunities to enhance ITS learning. For example, novel traffic signal control technologies based on reinforcement learning (RL), which learn adaptive signaling policies from simulations generated using real-world traffic data, have already achieved performance on par with and even exceeding traditional control methods \shortcite{Chen2020}.

However, collecting data for ITS learning remains a nontrivial task. First, many state-of-the-art deep learning approaches for ITSs inherently require large quantities of data for training \shortcite{Garg2019}, but they also need to be able to learn from a heterogeneous set of experiences so that they can achieve robust performance in the real world. Second, once they are trained, ITSs will need to be able to interact with many human users, including both end-users (i.e., drivers) and decision-makers (i.e., traffic engineers and city planners). End-users have unique needs and make unique choices that must be addressed individually and equitably. Decision-makers also have complex and potentially conflicting requirements \shortcite{Goh2012} that necessitate repeated iteration of the design of ITSs. Thus, data collection for ITS learning must be large-scale and continuously-occurring. However, performing data collection from real-world transportation systems in this manner may result in significant efficiency and safety costs \shortcite{Garg2019}.

Due to these issues, traffic simulators are commonly used to substitute or complement real-world data collection. They serve as safe sandboxes that can generate realistic data for both ITS learning \shortcite{Barcelo2004} and real-world decision-making \shortcite{AlvarezLopez2018,Fries2007}. This work narrowly focuses on open-source traffic simulators that have been developed for the training of ITSs based on RL methods, including systems for traffic signal control, autonomous driving, ramp metering, and tolling \shortcite{Haydari2022}. RL algorithms are particularly demanding in terms of the number of environmental interactions that they require, so traffic simulators for RL applications must be able to quickly generate high-fidelity, high-granularity data \shortcite{Zhang2019}. 

Released in 2001, SUMO \shortcite{AlvarezLopez2018} is the traffic simulator that has been most commonly used to train deep RL-based ITSs \shortcite{Noaeen2022}. It supports an expansive framework for simulation definitions, as well as efficient programmatic API access. However, it is single-threaded and thus scales less well to very large road networks. Motivated by this limitation, \shortciteN{Zhang2019} introduced CityFlow, a traffic simulator designed for applications of RL to traffic signal control. It is multi-threaded, and consequently was reported to achieve a speedup of $>$20x over SUMO. However, its framework for simulation definitions is more restricted and focuses on aspects which are essential for RL training. Its adoption is limited but increasing \shortcite{Noaeen2022}.

If traffic simulators are to serve as training environments for RL algorithms, their modelling assumptions must be sufficiently realistic that RL-based ITSs can learn to adapt to a variety of scenarios during deployment. Thus, real-world validation is crucial. However, granular validation with real-world data is usually not possible due to the aforementioned challenges of data collection. Comparisons between simulators take a partial step towards this goal by verifying that different simulators lead to equivalent outcomes. This work compares CityFlow against the more granular SUMO. \shortciteN{Zhang2019} included a preliminary comparison of CityFlow to SUMO using the expected travel time of vehicles in a road network. However, travel time is a long-term, system-level outcome, whereas RL algorithms use instantaneous, local features for learning. A more comprehensive comparison was performed to answer the following research questions: 
\begin{enumerate}[RQ1, wide=0pt, leftmargin=*]
    \item Do the low-level simulation outcomes of CityFlow and SUMO have a statistically significant level of distributional equivalence? 
    \item How is this distributional equivalence affected by incorporating variation between vehicles in terms of different driver behavioral models?
    \item How is this distributional equivalence affected by the scale of the simulation, in terms of traffic demand and road network size?
\end{enumerate}

\section{Related Work}
\label{sec:related}
\subsection{Validating Traffic Simulators}
\label{sec:related:traffic-sims}
Validation is an important yet challenging aspect of the development of traffic simulators that ensures their fidelity to the real world. A multitude of road networks have been used to validate SUMO itself \shortcite{Bedogni2015,AlvarezLopez2018} and to calibrate its car-following models \shortcite{Krajzewicz2002} in comparison to detector data. For instance, \shortciteN{Lobo2020} compared traces for vehicle counts and crossing times between SUMO and detector data for a road network in Ingolstadt, Germany; their simulation is used in this work. Meanwhile, CityFlow was validated by comparing average travel times under various traffic volumes to that of SUMO \shortcite{Zhang2019}.

A number of studies have compared outcomes from multiple simulators; this work falls in this setting. \shortciteN{Maciejewski2010} compared vehicle counts under different traffic demand and driver behavior settings for SUMO, the commercial simulator VISSIM, and TRANSIMS. Several studies involved SUMO and the commercial simulator AIMSUN. \shortciteN{Leksono2012} compared travel times and queue lengths in these simulators under different traffic management interventions for a roundabout in Norrk\"oping, Sweden. \shortciteN{Ronaldo2012} used $t$-tests to compare flows and speeds in these simulators to observations from a highway interchange in Stockholm, Sweden. \shortciteN{BazaSolares2022} applied $t$-tests to vehicle counts from the two simulators for a road network from Bucaramanga, Colombia. This work differs from prior approaches in that: (1) the measures used relate to the distribution of outcomes across the network, not just at individual points; and (2) these measures were evaluated across multiple road network scales. 

\subsection{Modelling Driver Behavior}
\label{sec:related:behavior}
There has been a significant body of literature on building realistic models of driver behavior. Car-following behavior was the first to be modelled, with the early Gazis-Herman-Rothery model being over 50 years old. Subsequent work has yielded optimal velocity, fuzzy logic, collision avoidance, action point, and cellular automaton models \shortcite{Li2012,Saifuzzaman2014}. Collision avoidance and action point models were implemented for this work. Lane-changing behavior has been modelled by a newer, separate line of work \shortcite{Moridpour2010a,Zheng2014}, which has produced models based on rules, discrete choice, game theory, and cellular automata. Some work has built unified models of car-following, lane-changing, and other driver behavior \shortcite{Toledo2007,Markkula2012}. Here, car-following and lane-changing models were considered separately but co-varied in virtual experiments to elucidate the effects of their interactions. The most relevant prior work is \shortciteN{CapelaDias2013}, who analyzed the impact of driver behavior on system-level travel time in SUMO; by contrast, this work focuses on distributional equivalence between two simulators in terms of lower-level outcomes.

\section{Traffic Simulators for RL}
\label{sec:traffic-sims}
This section reviews the use of traffic simulators for RL-based ITS training.

\subsection{RL for Transportation}
\label{sec:traffic-sims:rl}
RL algorithms learn policies to maximize their expected cumulative reward through repeated interactions with an environment, which for ITSs is the traffic simulator. Examples of cumulative reward objectives include travel time minimization for traffic signal control and speed limit control \shortcite{Noaeen2022,Wu2020}, and revenue maximization for tolling \shortcite{Pandey2020}. However, cumulative rewards are not useful for learning because they are system-level measures calculated over longer timespans. Instead, RL algorithms observe instantaneous vehicle-level state features, use them to select actions, and then receive instantaneous rewards \shortcite{Haydari2022}. 

Formulations of the observation and instantaneous reward spaces vary between papers, but they are rarely aligned directly with the cumulative reward. Observations usually consist of counts (e.g., queue lengths), positions, speeds, or other aggregated properties of individual vehicles \shortcite{Noaeen2022,Pandey2020,Wu2020}. Rewards are more problem-dependent. For traffic signal control, queue length, waiting time, and speed are common factors \shortcite{Koohy2022}; speed limit control may use proxies for travel time, crash probability, or emissions \shortcite{Wu2020}; and tolling may optimize a combination of vehicle count, travel time on segments, and collected tolls \shortcite{Pandey2020}. Regardless of these different formulations, it is evident that system-level measures involving the movement of vehicles through the whole road network are not directly used for RL training.

\subsection{Comparing CityFlow and SUMO}
\label{sec:traffic-sims:docking}
Microscopic traffic simulators are most often used for RL in transportation \shortcite{Haydari2022}. In contrast to macroscopic and mesoscopic simulators, microscopic simulators are primarily agent-based, as they model the behavior of individual vehicles \shortcite{Barcelo2010}. This is helpful for gathering the types of vehicle-level observations introduced in \autoref{sec:traffic-sims:rl}. Both CityFlow and SUMO are microscopic. They represent road networks as graphs, with intersections as nodes (``junctions'' in SUMO) and roads as edges between nodes. Vehicles are generated by flows; all vehicles within a flow share similar attributes, including behavioral parameters and routes. Routes are defined as sequences of edges. Where edges are joined at an intersection, vehicles select a pair of lanes (``roadlinks'' in CityFlow, ``connections'' in SUMO) to traverse the intersection. However, key differences exist in finer details. Unlike CityFlow, SUMO supports definitions for more complex types of lanes (e.g., sidewalks, bicycle lanes, and ramps) and vehicle classes (e.g., public transit and emergency vehicles, pedestrians, and cyclists). It also supports stochastic generation of routes and vehicle types, and more sophisticated traffic signalling, including yellow lights.

Where possible, the virtual experiments in this work were designed to control or co-vary differences between the simulators. In particular, flows in both CityFlow and SUMO were converted to be single-vehicle and deterministic, and to share the same routes. However, differences remain, particularly aspects of driver behavior (\autoref{sec:behavior}) driven by randomness in SUMO; by contrast, CityFlow only uses random generation for vehicle priority. Such uncontrolled factors may account for their differences.

\section{Varying Driver Behavior}
\label{sec:behavior}
Addressing RQ2 requires distributional equivalence to be assessed under different variations of driver behavior; as established in \autoref{sec:related:behavior}, these are implemented through car-following and lane-changing models in traffic simulators. In both CityFlow and SUMO, car-following logic is used to maintain a safe gap to the leading vehicle, while lane changes are used to switch vehicles between lanes so that they can take appropriate roadlinks at intersections to continue their routes. However, in CityFlow, car-following and lane-changing are handled in separate threads; in SUMO, they are handled in sequence, with car-following logic being executed before lane-changing logic.

\subsection{Car-Following Models}
\label{sec:behavior:car-following}
Car-following models determine the speeds at which vehicles travel unobstructed (free speed), follow behind a lead vehicle (following speed), and stop at an obstacle (stopping speed). Two types of numerical integration can be used to calculate vehicle speeds: a Euler update, which solves for the speed at discrete timesteps, and a ballistic update, which solves for the acceleration at discrete timesteps and applies it to the speed. Both simulators were controlled to use ballistic updates.

\begin{table}[H]
    \centering
    \caption{Parameter settings for six aggressiveness types based on Capela Dias et al. (2013). Maximum emergency deceleration was uniformly set to $-9.0 \mathrm{\sfrac{m}{s^2}}$, as they did not specify this parameter.}
    \begin{tabulary}{\textwidth}{LCCCCC}
        \toprule
        \textbf{Type} & \textbf{Max. accel.} & \textbf{Max. decel.} & \textbf{Max. emerg. decel.} & \textbf{Min. gap} & \textbf{Min. headway} \\
        \midrule
        Aggressive young & $3.1\ \mathrm{\sfrac{m}{s^2}}$ & $-5.5\ \mathrm{\sfrac{m}{s^2}}$ & $-9.0\ \mathrm{\sfrac{m}{s^2}}$ & 1.2 m & 1.0 s \\
        Courteous young & $2.5\ \mathrm{\sfrac{m}{s^2}}$ & $-4.5\ \mathrm{\sfrac{m}{s^2}}$ & $-9.0\ \mathrm{\sfrac{m}{s^2}}$ & 2.5 m & 1.0 s \\
        \shortstack[l]{Aggressive \\ middle-aged} & $2.9\ \mathrm{\sfrac{m}{s^2}}$ & $-5.0\ \mathrm{\sfrac{m}{s^2}}$ & $-9.0\ \mathrm{\sfrac{m}{s^2}}$ & 2.0 m & 1.3 s \\
        \shortstack[l]{Courteous \\ middle-aged} & $2.4\ \mathrm{\sfrac{m}{s^2}}$ & $-4.1\ \mathrm{\sfrac{m}{s^2}}$ & $-9.0\ \mathrm{\sfrac{m}{s^2}}$ & 2.5 m & 1.5 s \\
        Aggressive old & $2.6\ \mathrm{\sfrac{m}{s^2}}$ & $-4.5\ \mathrm{\sfrac{m}{s^2}}$ & $-9.0\ \mathrm{\sfrac{m}{s^2}}$ & 2.0 m & 1.7 s \\
        Courteous old & $2.3\ \mathrm{\sfrac{m}{s^2}}$ & $-3.8\ \mathrm{\sfrac{m}{s^2}}$ & $-9.0\ \mathrm{\sfrac{m}{s^2}}$ & 2.5 m & 1.9 s \\
        \bottomrule
    \end{tabulary}
    \label{tab:aggressiveness}
\end{table}

Several shared parameters have varying effects on different car-following models. CityFlow assumes that vehicles have usual and maximum accelerations and decelerations; SUMO assumes that vehicles have a maximum possible acceleration and deceleration (which are used as the usual values), and a maximum emergency deceleration. The latter formulation was used here, although this may have resulted in more aggressive behavior than if lower usual accelerations and decelerations were used. Additionally, both simulators model vehicles as having minimum desired following distances in terms of space (i.e., the minimum gap) and time (i.e., the minimum headway). \shortciteN{CapelaDias2013} co-varied these parameters to model the effect of driver aggressiveness on travel time. This work adopts their taxonomy of aggressiveness types, but excludes gender effects due to conflicting conclusions regarding their significance in the literature \shortcite{TaiebMaimon2001,Tasca2000}. This gave six parameter settings (\autoref{tab:aggressiveness}).

The default car-following models in CityFlow and SUMO are both modified from the collision avoidance model of \shortciteN{Krauss1998}. For free speed, CityFlow's implementation uses the maximum speed, while SUMO's implementation modulates this by the visible lookahead distance. For stopping speed, both simulators solve somewhat different quadratic equations to determine the deceleration needed to stop within a fixed distance. For following speed, a target distance is maintained, which is computed by the desired minimum headway as well as the speed and maximum deceleration of the lead vehicle. In this work, SUMO's variant of the Krau{\ss} model was added to CityFlow. SUMO implements several other car-following models, which were also re-implemented in CityFlow to introduce variation in driver behavior:
\begin{itemize}
    \item The collision avoidance/action point model of \shortciteN{Wagner2008}, which probabilistically combines Krau{\ss}'s model with the action point model of \shortciteN{Todosiev1963}.
    \item The collision avoidance model of \shortciteN{Wiedemann1974}, which is a behavioral model that varies between free, approaching, following, and emergency modes based on the gap to the lead vehicle.
    \item The adaptive cruise control (ACC) model of \shortciteN{Milanes2014}, which determines acceleration using a ``speed control'' method if the gap to the lead vehicle is large and using a ``gap control'' method if the gap is small.
\end{itemize}

\subsection{Lane-Changing Models}
\label{sec:behavior:lane-changing}
In both CityFlow and SUMO, lane changes are initiated by signaling the lead and lag vehicles on the destination lane, and the vehicle is instantly moved to the other lane upon completion of the lane change. (SUMO also supports a sublane model that explicitly models lateral movement, which was excluded as a control.) However, the implementation details of lane-changing likewise differ between these simulators. Vehicles in CityFlow only perform lane changes to follow their routes, and lane-changing is based on the explicit insertion of a copy of the vehicle (the ``shadow vehicle'') on the target lane that the lag vehicle will follow. Vehicles in SUMO follow a parameterized hierarchy of motivations, which includes strategic lane changes to follow routes as well as tactical lane changes for overtaking. 

One of SUMO's parameters, the gap tolerance factor, was introduced to CityFlow in this work. When vehicles are changing lanes, the minimum gap on the destination lane required for a vehicle to initiate a lane change is given by the necessary gap for collision avoidance divided by this constant factor. A factor of 1.0 represents the default behavior. The virtual experiments in this work varied this factor between 0.5, 0.82, 1.0, 1.18, and 1.5, representing decreases/increases of 18\% and 50\% in tolerance. This is based on \shortciteN{Sun2010}, who observed the mean gaps for forced, cooperative, and free lane change maneuvers to be 45 ft, 53 ft, and 109 ft in a video dataset.

\section{Experimental Setup}
\label{sec:exp-setup}
To address the research questions, two virtual experiments were designed. Both address RQ1 and RQ2 by assessing CityFlow and SUMO's distributional equivalence under different settings of driver behavior. The two experiments also address separate aspects of RQ3: Experiment 1 assesses distributional equivalence while varying the simulation scale by traffic demand, and Experiment 2 while varying it by road network size. This was accomplished using road networks from the benchmark dataset of \shortciteN{Ault2021}, as described in the following subsections. All of these networks were initially defined using SUMO syntax; the SUMO-to-CityFlow network converter of \shortciteN{Zhang2019} was used to generate their CityFlow counterparts, and a flow converter was created to map between vehicular flows in the two simulators.

Four independent variables were considered in both virtual experiments: the car-following model (\autoref{sec:behavior:car-following}, 5 levels), the car-following aggressiveness parameters (\autoref{sec:behavior:car-following}, 6 levels), the lane-changing gap tolerance (\autoref{sec:behavior:lane-changing}, 5 levels), and the road network (2 levels). For the car-following model, the default and SUMO-based implementations of the Krau{\ss} model in CityFlow were compared to Krau{\ss} model in SUMO, while the other models' CityFlow implementations were compared to their respective SUMO counterparts. The fundamental lane-changing models for the two simulators were held constant, as was the traffic signal program --- a simple fixed-time program retained from the original dataset.

Eight instantiations of two types of measures were used to assess distributional equivalence. First, the root mean squared error (RMSE) quantifies the point-to-point difference in individual outcome measures. This was computed as the mean RMSE of the total travel time and waiting time (defined as time that a vehicle spends queued with a speed $< 0.1 \mathrm{\sfrac{m}{s}}$) over all vehicles and timesteps; the mean RMSE of per-lane total vehicle counts and queued vehicle counts over all lanes and timesteps; and the mean RMSE of the speed and acceleration over all individual vehicles and timesteps. Second, the Kullback-Liebler (KL) divergence measures the difference in the distribution of outcomes over the entire road network. This was computed for the distributions of vehicle counts and queued vehicle counts as the mean over all timesteps. 

\subsection{Experiment 1: Traffic Demand}
\label{sec:exp-setup:exp-1}

\begin{figure*}[ht]
\centering
\begin{subfigure}{0.4\textwidth}
    \centering
    \includegraphics[width=\textwidth]{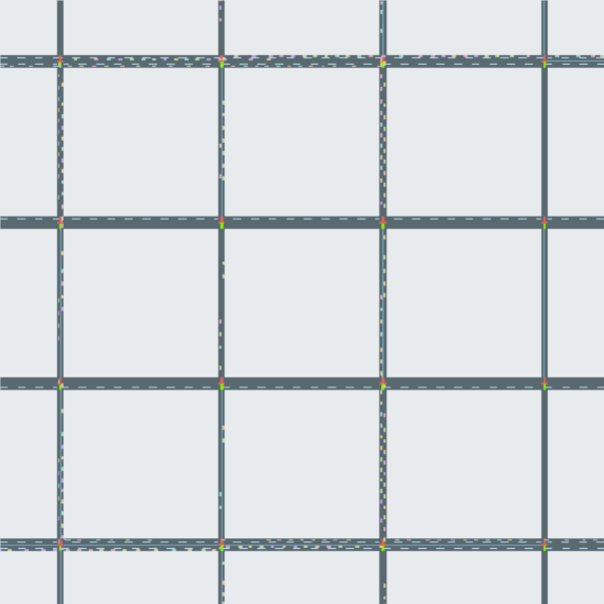}
    \caption{\textbf{arterial4x4}}
    \label{fig:arterial4x4}
\end{subfigure}
\hspace{5em}
\begin{subfigure}{0.4\textwidth}
    \centering
    \includegraphics[width=\textwidth]{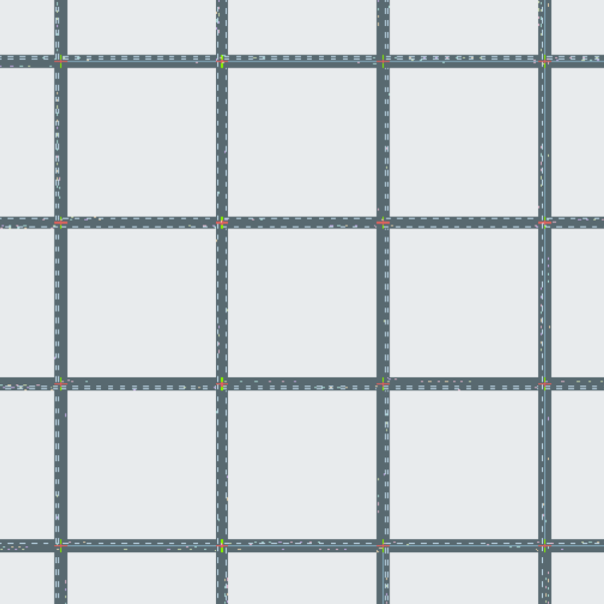}
    \caption{\textbf{grid4x4}}
    \label{fig:grid4x4}
\end{subfigure}
\caption{Screenshots in CityFlow of the arterial4x4 and grid4x4 road networks.}
\label{fig:grid}
\end{figure*}

For this experiment, the road networks arterial4x4 \shortcite{Ma2020} and grid4x4 \shortcite{Chen2020} (\autoref{fig:grid}) from \shortciteN{Ault2021}'s RESCO benchmark were used. Both are synthetic grid networks with similar topologies, but they vary in the level of congestion. grid4x4 consists of six-lane roads with a uniformly distributed demand of 1,473 vehicles. arterial4x4 consists of major (four-lane) and minor (two-lane) roads, and a demand of 2,484 vehicles that alternates between major and minor roads. arterial4x4's traffic pattern leads to congestion and degraded RL performance in simulations \shortcite{Ault2021}.

The following power analysis uses these variable names: $C$ for car-following models, $A$ for car-following aggressiveness, $L$ for lane-changing gap tolerance, and $R$ for road network. The second-order linear multiple regression included 5 ($C$) + 6 ($A$) + 1 ($L$) + 1 ($L^2$) + 2 ($R$) + 10 ($C \cdot R$) + 12 ($A \cdot R$) + 2 ($L \cdot R$) + 30 ($C \cdot A$) + 5 ($C \cdot L$) + 6 ($A \cdot L$) = 80 variables. Using G*Power 3.1.9.7's power calculation for ordinary linear multiple regression with a fixed model and $R^2$ increase, a small effect size of 0.02, and $\alpha = \beta = 0.95$, the total necessary sample size was computed as 2,646. Divided by the number of cells, $5 \cdot 6 \cdot 5 \cdot 2 = 300$, the number of replications per cell was computed as $\lceil\frac{2,646}{300}\rceil = 9$. All replications were executed using Python 3.9.16, SUMO 1.12.0, and a modified version of CityFlow 0.1 on a shared server with four cores, two 4.2GHz Intel i7-7700K processors per core, and 62 GiB of RAM.

\subsection{Experiment 2: Network Scale}
\begin{figure*}[ht]
\centering
\begin{subfigure}{0.4\textwidth}
    \centering
    \includegraphics[width=\textwidth]{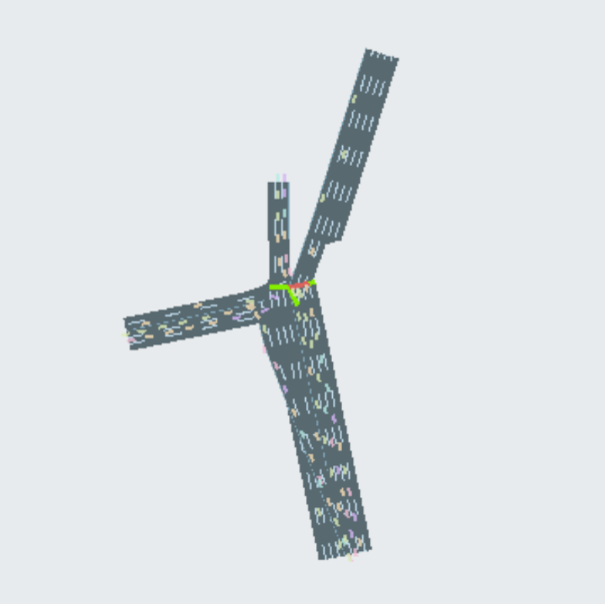}
    \caption{\textbf{ingolstadt1}}
    \label{fig:ingolstadt1}
\end{subfigure}
\hspace{5em}
\begin{subfigure}{0.4\textwidth}
    \centering
    \includegraphics[width=\textwidth]{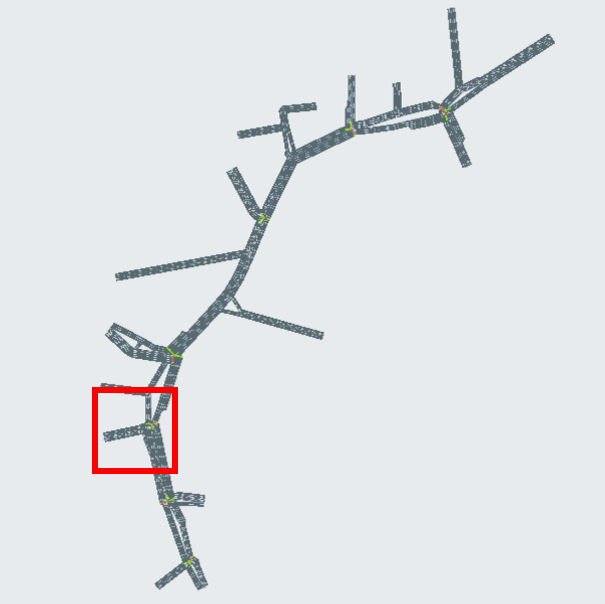}
    \caption{\textbf{ingolstadt7}}
    \label{fig:ingolstadt7}
\end{subfigure}
\caption{Screenshots in CityFlow of the ingolstadt1 and ingolstadt7 road networks.}
\label{fig:ingolstadt}
\end{figure*}

For this experiment, the road networks ingolstadt1 and ingolstadt7 (\autoref{fig:ingolstadt}) from \shortciteN{Ault2021}'s RESCO benchmark were used. Both are subsets of the Ingolstadt road network that was simulated by \shortciteN{Lobo2020}. They respectively contain 1 and 7 signalized intersections, representing a single busy intersection and a larger arterial road; ingolstadt7 is a superset of ingolstadt1. The total demands of the two road networks are respectively 1,716 and 3,031 vehicles.

\label{sec:exp-setup:exp-2}

Based on the analysis in \autoref{sec:exp-setup:exp-1}, the number of replications per cell was computed as $\lceil\frac{2,646}{300}\rceil = 9$. All replications were executed using Python 3.9.16, SUMO 1.12.0, and a modified version of CityFlow 0.1 on a shared server with four cores, two 4.2GHz Intel i7-7700K processors per core, and 62 GiB of RAM.

\section{Experimental Results}
\label{sec:exp-results}
 One-sample $t$-tests indicated that all of the RMSE and KL divergence measures were significantly different from 0, with a $p$-value $\ll 0.001$ for all cells in Experiments 1 and 2. This suggests a lack of distributional equivalence between CityFlow and SUMO. The following subsections explore the results of second-order linear multiple regression for each of the measures. Notably, coefficients involving the difference between the SUMO and CityFlow Krau{\ss} model implementations were generally not significant.

\subsection{Experiment 1: Traffic Demand}
\label{sec:exp-results:exp-1}
For total time and waiting time RMSEs, the road network, car-following model, and aggressiveness generally had significant effects, as did various pairwise interactions between them. The uncongested grid4x4 network had significantly lower RMSEs (coefficients: $-890.61$/$-1035.46$) than the congested arterial4x4 network (intercepts: $1718.61$/$1815.46$), suggesting that congestion worsened the distributional equivalence of these measures. In arterial4x4, the Wagner (coefficients relative to SUMO Krau{\ss}: $546.8$/$466.65$) and Wiedemann models (coefficients: $100.96$/$208.99$) had significantly higher RMSEs; these differences were smaller for grid4x4 (coefficients relative to SUMO Krau{\ss}: $49.5$/$33.75$ for Wagner; $48.91$/$34.85$ for Wiedemann).

For total and queued vehicle count RMSEs and KL divergences, the road network, car-following model, aggressiveness, and gap tolerance generally had significant effects, as did various pairwise interactions between them. The RMSEs and KL divergences showed distinct patterns: the RMSEs were much lower for grid4x4 than arterial4x4 (coefficients: $-5.15$/$-4.93$), but the KL divergences had less variation (coefficients: $-0.621$/$-0.013$). Yet, the KL divergences had low enough standard deviations that the road network's effects remained significant. High aggressiveness in arterial4x4 generally yielded higher RMSEs (coefficients for aggressive young relative to aggressive middle-aged: $2.16$/$2.46$) and KL divergences (coefficients: $0.228$/$0.139$). While the same was true for the RMSEs in grid4x4, its KL divergences were lower for more aggressive settings (coefficients: $-0.204$/$-0.122$). Despite higher time measures, the Wagner model led to lower measures for total vehicle count (coefficients relative to SUMO Krau{\ss}: $-1.16$/$-0.193$).

For vehicle speed and acceleration RMSEs, the road network, car-following model, and aggressiveness generally had significant effects, as did various pairwise interactions between them. Greater equivalence in vehicle distributions did not always correspond to more similar vehicle-level measures. Both speed and acceleration RMSEs increased for grid4x4 (coefficients: $2.62$/$0.418$) even though the other measures were lower on average. Also, unlike its vehicle count measures but like its time measures, the Wagner model had significantly higher speed and acceleration RMSEs (coefficients relative to SUMO Krau{\ss}: $2.71$/$1.26$).

\subsection{Experiment 2: Network Scale}
\label{sec:exp-results:exp-2}
For total time and waiting time RMSEs, the aggressiveness and its interactions generally had significant effects, along with the Wiedemann model and its interactions. The best-fitting model for total time did not include a road network-gap tolerance interaction, whereas the model for waiting time did. The smaller ingolstadt1 network had lower but more variable RMSEs (intercepts: $899.29$/$1170.3$), while the larger ingolstadt7 network had significantly higher but more uniform RMSEs (coefficients: $1264.59$/$1046.07$). For ingolstadt1, the most aggressive parameter settings led to significantly higher RMSEs (coefficients of aggressive young relative to aggressive middle-aged: $408.36$/$568.65$). Likewise, the Wiedemann model had significantly higher RMSEs than other car-following models in ingolstadt1 (coefficients relative to SUMO Krau{\ss}: $192.29$/$1917.31$), but this effect was reversed for ingolstadt7 (coefficients: $-349.13$/$-464.13$).

For total and queued vehicle count RMSEs and KL divergences, the road network and aggressiveness generally had significant effects, as did various pairwise interactions between them and with the gap tolerance. Unlike Experiment 1, the RMSE measures were lower in ingolstadt7 than in ingolstadt1 (coefficients: $-2$/$-1.67$), but the KL divergence measures were higher (coefficients: $0.461$/$2.48$). More aggressive parameter settings again led to significant increases in the RMSEs and KL divergences, with a larger increase in RMSEs (coefficients of aggressive young relative to aggressive middle-aged: $1.35$/$2.29$) than in KL divergences (coefficients: $0.595$/$0.717$). However, the increase in both measures was smaller in ingolstadt7 (coefficients: $0.764$/$1.507$ for RMSE, $0.283$/$0.413$ for KL divergences). Both the Wagner and ACC models had RMSEs that significantly increased with gap tolerance (coefficients relative to SUMO Krau{\ss} per unit of gap tolerance: $0.151$/$0.056$ for Wagner, $0.165$/$0.092$ for ACC), but KL divergences that significantly decreased with it (coefficients: $-0.285$/$-0.511$ for Wagner, $-0.354$/$-0.642$ for ACC). 

For vehicle speed and acceleration RMSEs, the Wagner and Wiedemann models along with the aggressiveness had significant effects, as did various pairwise interactions of the road network with the car-following models and aggressiveness. Again, the road network had significant effects on both speed and acceleration RMSEs, with these measures being higher for ingolstadt7 (coefficients: $5.51$/$0.173$). The Wagner car-following model had significantly higher RMSEs (coefficients relative to SUMO Krau{\ss}: $0.363$/$0.461$), whereas the Wiedemann model had significantly lower RMSEs (coefficients: $-1.94$/$-0.957$).

\subsection{Parameter Validity}
\label{sec:exp-results:validation}
Parameter validation was conducted by comparing the parameter settings for car-following and lane-changing models used in the virtual experiments to prior driving simulator and real-world studies. Overall, the dependence of these parameters on external factors such as traffic density and speed suggests that the settings used in traffic simulators should be calibrated to specific road networks and conditions. However, the settings used in this work remain largely reasonable considering the variation reported in the literature.

For acceleration and deceleration, the settings used in the virtual experiments were based on \shortciteN{CapelaDias2013}, with smaller values for older and less aggressive drivers (\autoref{sec:behavior:car-following}); however, they considered gender effects to be negligible. Similar values have been reported in prior work \shortcite{Nakagawa2006,Fadhloun2015,Korber2016}. However, among driving simulator studies, \shortciteN{Korber2016} demonstrated an age effect opposite to that assumed by Capela Dias et al. Among real-world studies, \shortciteN{Moon2008} found a dependence of the 95th percentile of braking decelerations on speed, and \shortciteN{Nakagawa2006} reported a significant interaction between age and gender.

For minimum gap and headway time, the settings used in the virtual experiments also followed \shortciteN{CapelaDias2013}, with larger values for older and less aggressive drivers (\autoref{sec:behavior:car-following}). Similar values have again been reported in prior work \shortcite{Michael2000,Vogel2002}. In particular, real-world studies for which reported values are influenced by a similar age effect include \shortciteN{Delorme2001,TaiebMaimon2001}, and \shortciteN{Moon2008}. \shortciteN{TaiebMaimon2001} and \shortciteN{Shangguan2019} reported larger minimum gaps and smaller minimum headways, but their measurements were made for higher speeds (respectively over $50\ \mathrm{\sfrac{km}{h}}$ and $100\ \mathrm{\sfrac{km}{h}}$).

For lane-changing gap tolerance, this work used gap widths in empirical lane-changing behavior to approximately quantify its level of variation, specifically using the mean gaps for forced, cooperative, and free lane changes from \shortciteN{Sun2010}'s real-world study (\autoref{sec:behavior:lane-changing}). The settings in this work are consistent with standard deviations in lead and lag gaps as reported in prior studies \shortcite{Moridpour2008,Balal2014}. \shortciteN{Ali2020}'s driving simulator study identified significant factors that impact gap tolerance: relative to the average male middle-aged driver, gaps are smaller for younger drivers, larger for female drivers, and smaller as speed increases. Likewise, \shortciteN{Hill2015}'s real-world study reported that the mean and standard deviation of lag gaps depended on congestion. Future work could use these factors to create a taxonomy of lane-changing behavior similar to \shortciteN{CapelaDias2013}.

\section{Conclusion}
\label{sec:conclusion}
In this work, virtual experiments compared the low-level simulation outcomes of two traffic simulators, CityFlow and SUMO. To capture the effects of modelling real-world heterogeneity, various parameters of driver behavior and road network scale were varied. The results indicate a lack of distributional equivalence between the simulators, with certain parameter settings worsening distributional equivalence.

However, as noted in \autoref{sec:traffic-sims}, this work is insufficient to provide a complete characterization of what the critical differences between these simulators are. Many aspects of CityFlow and SUMO that were not controlled --- simulation control-flow, other aspects of driver behavior, the effects of traffic signals, and randomness --- could all have contributed to the observed discrepancies. Thus, future work should perform more comprehensive, controlled evaluations of these two simulators. 

Regardless, researchers in RL for transportation must not take traffic simulators for granted as a deus ex machina for training, and must recognize that they may not be interchangeable. Which simulator, then, should be chosen? This work does not aim to answer this question, but some observations can be made:
\begin{itemize}
    \item \textbf{SUMO} provides a detailed simulation that models real-world heterogeneity, and captures additional aspects of traffic management and driver behavior
    \item \textbf{CityFlow} provides an efficient simulation that abstracts out and homogenizes various details, reducing the number of parameters that need to be tuned
\end{itemize}
Therefore, as with many other problems in simulation, the core trade-off between these two simulators (and others) involves veridicality and efficiency. There is no one best simulator; researchers must decide whether using a coarser abstraction of the environment is acceptable in exchange for faster training. 

But how exactly should researchers make this decision? Crucially, RL-based ITSs may not necessarily perform better when they are trained with more granular simulators. Prior work has shown that both introducing unnecessary complexity \shortcite{Zheng2019} and removing needed complexity \shortcite{Genders2018} in the observation space may harm performance. Researchers should compare training results from different simulators and use them to design RL formulations in a principled way. As a first step, \shortciteN{Goel2023} report training results on both SUMO and CityFlow, but they do not use the same road networks for both simulators. One strategy may be to train a baseline using an efficient simulator (e.g., CityFlow), and then to finetune it by further training with a veridical one (e.g., SUMO).

Ultimately, the true goal is to ensure that RL-based ITSs can perform well under real-world traffic conditions, which requires that their training and validation environments are close to reality. As previously noted, distributional equivalence between simulators is a proxy measure of this goal. Unfortunately, a chicken-and-egg problem exists in that traffic simulators are intended to replace real-world data collection, yet cannot be validated without it. For now, simulations should still be developed in collaboration with stakeholders to ensure that they meet acceptable standards of fidelity. However, the future holds promise for traffic simulations that are both veridical and efficient: the increasing prevalence of connected vehicles \shortcite{Mathew2021} means that collection of granular real-world data for validation may be within reach.

\section*{Acknowledgments}
This work was supported by a Mobility21 research grant and the Tang Family Endowed Innovation Fund.

% Reducing font size (to 9pt) for References & Author Biographies
\footnotesize

% Please don't exchange the bibliographystyle style
\bibliographystyle{wsc}
\bibliography{ref}

\section*{Author Biographies}
\label{sec:appendix:bios}

\noindent \textbf{REX CHEN} (email \email{rexc@cmu.edu}) is a third-year PhD candidate in the Societal Computing program of the Software and Societal Systems Department in the School of Computer Science at Carnegie Mellon University. His research focuses on applications of reinforcement learning and other AI techniques to socially impactful domains such as transportation. His work aims to evaluate the impact of real-world uncertainty and user behavior on AI algorithms designed for these domains.
\newline

\noindent \textbf{KATHLEEN M. CARLEY} (email \email{kathleen.carley@cs.cmu.edu}) (H.D. University of Zurich, Ph.D. Harvard, S.B. MIT) is a Professor of Societal Computing, in the School of Computer Science’s Software and Societal Systems Department at Carnegie Mellon University; and the Director of the Center for Computational Analysis of Social and Organizational Systems (CASOS), and the Center for Informed Democracy and Social Cybersecurity (IDeaS). She and her teams have developed infrastructure tools for analyzing large-scale dynamic networks, bot hunters, and various agent-based simulation systems. 
\newline

\noindent \textbf{FEI FANG} (email \email{feifang@cmu.edu}) is the Leonardo Assistant Professor in the Software and Societal Systems Department in the School of Computer Science at Carnegie Mellon University. Her research interests lie in the area of artificial intelligence and multi-agent systems, focusing on the integration of computational game theory and machine learning to address real-world challenges in critical domains such as security, sustainability, and mobility.
\newline

\noindent \textbf{NORMAN SADEH} (email \email{sadeh@cs.cmu.edu}) is a professor in the School of Computer Science at CMU, where he is affiliated with the Software and Societal Systems Department, the Human-Computer Interaction Institute, and the CyLab Security and Privacy Institute. His research interests span mobile computing, the IoT, cybersecurity, privacy, machine learning, AI, and related public policy issues. His past work includes deployed planning and scheduling technologies for commercial systems.
\end{document}